# Multi-class Active Learning: A Hybrid Informative and Representative Criterion Inspired Approach

Xi Fang, Zengmao Wang, Xinyao Tang, Chen Wu

*School of Computer, Wuhan University, Wuhan, China*

(xifang96@qq.com, zengmaowang@whu.edu.cn, 403609158@qq.com, chen.wu@whu.edu.cn )

*Abstract*--- Labeling each instance in a large dataset is extremely labor- and time- consuming . One way to alleviate this problem is active learning, which aims to   which discover the most valuable instances for labeling to construct a powerful classifier. Considering both *informativeness and representativeness* provides a promising way to design a practical active learning.  However, most existing active learning methods select instances favoring either *informativeness* or *representativeness*. Meanwhile, many are designed based on the binary class, so that they may present suboptimal solutions on the datasets with multiple classes. In this paper, a hybrid informative and representative criterion based multi-class active learning approach is proposed. We combine the informative *informativeness* and *representativeness* into one formula, which can be solved under a unified framework. The informativeness is measured by the margin minimum while the representative information is measured by the maximum mean discrepancy. By minimizing the upper bound for the true risk, we generalize the empirical risk minimization principle to the active learning setting. Simultaneously, our proposed method makes full use of the label information, and the proposed active learning is designed based on multiple classes. So the proposed method is not suitable to the binary class but also the multiple classes. We conduct our experiments on twelve benchmark UCI data sets, and the experimental results demonstrate that the proposed method performs better than some state-of-the-art methods.

## I.    INTRODUCTION

However, in many real world problems, unlabeled data are easy to obtain, while labeling is usually expensive and time consuming due to the involvement of human experts. Hence, it is significance to train a good classifier with limited labeled data. Active learning addresses this challenge by querying the most informative samples to make the large amount of unlabeled data annotated automatically.

The key component of active learning methodologies depends on the design of an available criterion to select the most valuable samples iteratively. Two primary strategies, i.e. informativeness and representativeness, are widely used to design practical active learning algorithms [1]. The first one measures the ability of the unlabeled data in reducing the uncertainty of the current classifier, which could shrink the classifiers rapidly. The second strategy measures the representativeness of samples in the all input unlabeled data, it denotes whether an instance well represents the distribution of overall unlabeled data. Most active learning algorithms just employ one of the two strategies. The most exemplary approaches of querying informative samples for active learning include expected error reduction [2, 3], query by committee [4, 5, 6], the most uncertain criteria [1, 7, 8], the deficiency of such methods is that the queried instances could not exploit the structure of unlabeled data, and guarantee to be independent and identically distributed from the original data, since the queried instances solely depend on little labeled data [9]. When training the classifier with the informative samples, it will lead to serious sample bias and consequently undesirable performance [10]. The second strategy of active learning algorithms aims to query representative samples for the overall patterns of unlabeled data [1, 11]. Such type methods can perform better when there is few or no initial labeled data. However, since lack of the uncertain information, they have to query relatively large number of unlabeled data before the optimal classifier is trained and the efficiency also will degrade with the queried data increasing.

Since deploying either kind of strategy will significantly limit their performance. Several works have been done trying to query samples with high informativeness and high representativeness [12, 13, 14, 15]. They are mostly ad hoc in measuring the informativeness and representativeness of an instance, leading to suboptimal performance. Recently, Li and Guo [16] try to apply informative and representative information together, they calculate the two terms respectively, and then the two terms combined with multiplication and balanced by a weight. They use the conditional entropy as the informative measurement, and use Gaussian Process framework to measure the representativeness of unlabeled data, in this process, an inverse matrix of covariance matrix whose size is equal to the length of unlabeled data should be calculated. However, if the unlabeled data are too few, the covariance matrix will be singular or near singular. If the unlabeled data are too large, the computation complexity will be increased rapidly, and especially, if the unlabeled data not satisfy the normal or near normal distribution, the good performance will not be achieved. In [10], Huang el at try to use both informative and representative information in one optimization formulation based on max-min view [20]. They use unlabeled data in the semi-supervised learning setting for boosting the learning performance. However, the queried samples may not preserve the original data distribution. If the data structure does not satisfy semi-supervised assumptions [18, 19], the performance would not be desired. In addition, the methods that are described above are all not consider the label information and designed based on binary class.

Here, we aim to make full use of the label information to boost the active learning performance. Inspired by [10] and [17], we present a novel active learning approach called *Multi-class Active Learning by Hybrid Informative and Representative Information* (McALHIRI), which addresses the following two objectives: 1) to query the instances contains the most informative information; 2) to enforce the query instances that are diverse with each other in the whole active learning procedure. Our main contributions include:

1) For the multiple classes, the label information of each instance is fully used. Generally, for the single label classification, for each instance, there is just a label which is positive or negative to be used in the method, the uncertainty is just measured by one binary classifier. In the proposed multi-class active learning, for an instance, the label is a vector which just contains one positive value corresponding to the class that the instance belongs to. In this situation, the query instance can be closest to all classification hyperplanes between binary classes.
2) We combine the informative and representative information together, and by minimizing the empirical risk, the true risk under the original data distribution is approximately minimized. Meanwhile, in the proposed formula, a non-convex problem exists. An algorithm is proposed to solve it.
3) The proposed method can suffer from little initial labeled data issue, which might result slow convergence for active learning. We adopt the MMD to select the most representative information instances to meet the challenge that the discriminative information is lack, and thus to make sure boost the active learning performance.

We have demonstrated the proposed algorithm on several UCI benchmark data sets. Results indicate that our proposed algorithm is superior in comparison with the state-of-the-art active learning methods.

The rest of this paper is organized as follows. Section II provides the background of the related knowledge about active learning and correlation theory to the proposed method. In section III, we introduce our proposed method in details, and following we describe the experiments exhaustively to demonstrate the efficiency of our method in section IV. Finally, a simple summarization about our method and a brief discussion for future research directions are mad.

## II. PRELIMINARY

In the proposed framework, we derive an empirical risk for active learning risk and adapt maximum mean discrepancy to measure the representative information. Minimizing empirical risk has been successfully applied in machine learning and data mining methods [10, 21, 22]. In [22, 23], it has demonstrated that minimizing true risk under unseen data distribution is approximated by the summation of empirical risk on the labeled data and a properly designed regularization term, which constrains the complexity of the candidate classifiers.

### A. Active Learning

Active learning can be modeled as a quintuple ($T, F, U, Q, S$), it is an effectively approach to solve the small sample problem. In active learning, $T$ is the labeled data set with limited samples, $F$ is the classifier model trained by $T$, $U$ is the pool of samples which contains abundant samples that are unlabeled, Q is a data set with samples that query from U, the length of Q can be 1 or a batch, S is a superior that is busy in correctly labeling Q. As described above, active learning is an iteratively process, at each iteration, the query set Q is added into T, and removed from unlabeled set U, it stops until the classified model is robust or the samples reach a fixed number. In fig.1, it shows the process of active learning.

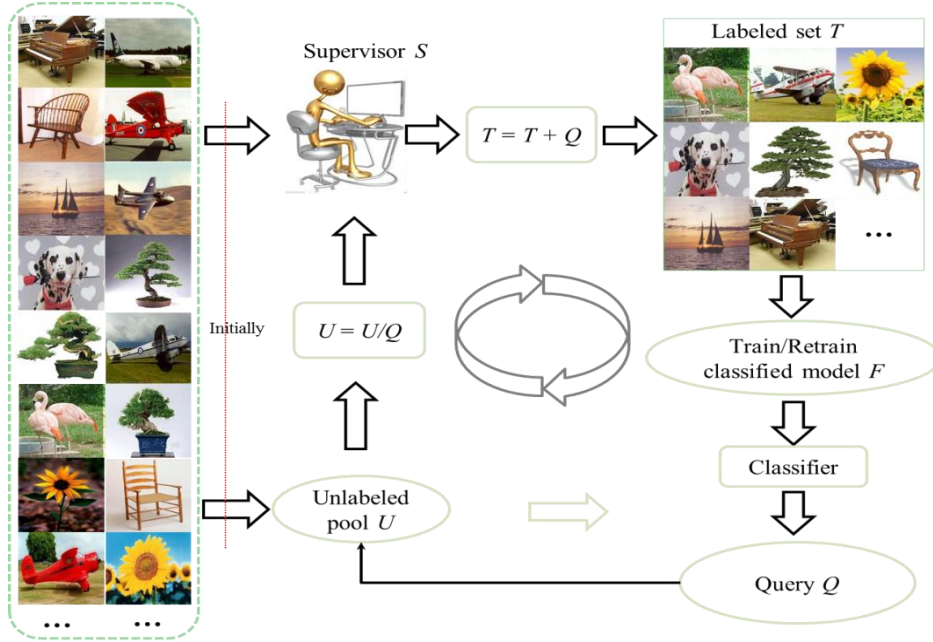

Fig.1 The flowchart of Active Learning

B.  Maximum Mean Discrepancy(MMD)

Assume $X = \{x_1, x_2, \ldots, x_n\}$ and $Y = \{y_1, y_2, \ldots, y_m\} \in R^d$ are two datasets drawn randomly from a source data. Let $p$, $q$ be two probability measures defined on $X$ and $Y$, respectively. The Maximum Mean Discrepancy is proposed to address the problem whether two probably distribution $p$ and $q$ are similar or not. The principle potential of the MMD is to find a function that assumes different expectations between different distributions so that when calculate the similar between two distributions it can evaluate empirically. So the quality of the MMD depends on the class functions. Let $\mathcal{F}$ be a class of functions $f : \chi \to R$, and let $p$, $q$, $x$, $y$, $X$, $Y$ be defined as above. According [24], the MMD is defined as:

$$MMD[F, p, q] = \sup_{f \in F} \left( E_p[f(x)] - E_q[f(y)] \right) \quad (1)$$

In the statistics [25], the empirical estimate of MMD can be replaced by the empirical expectation computed on the samples $X$ and $Y$, the empirical MMD can be defined as:

$$MMD[F, X, Y] = \sup_{f \in F} \left( \frac{1}{n} \sum_{i=1}^{n} f(x_i) - \frac{1}{m} \sum_{i=1}^{m} f(y_i) \right) \quad (2)$$

Clearly, when $\mathcal{F}$ is 'rich enough', the MMD will be vanish if and only if $p = q$, and if $\mathcal{F}$ is 'restrictive' enough, the empirical estimate of MMD will convergence quickly to its expectation since the data size increases. It has attested that the unit ball in reproducing kernel Hilbert space (RKHS) with a characteristic kernel can satisfy both of the foregoing properties [26], [27]. So the unit ball in characteristic RKHS can be used as the class of functions, then the MMD[F, X, Y] will be zero if and only if $p = q$. Let $\mathcal{H}$ be an RKHS, $\phi : X \to \mathcal{H}$ is known as feature space mapping from original to $\mathcal{H}$, $\mathcal{F}$ is a class of functions defined as the unit ball in characteristic RKHS, the MMD can be defined in RKHS, which can detect the all discrepancies between $X$ and $Y$ in characteristic RKHS. The empirical estimate of MMD in RKHS is defined as

$$MMD_\phi[X,Y] = \left\| \frac{1}{n}\sum_{i=1}^{n}\phi(x_i) - \frac{1}{m}\sum_{i=1}^{m}\phi(y_i) \right\|_{\mathcal{H}}^{2} \qquad (3)$$

### III. Multi-class Active Learning

Our approach is motivated by the multi-label classification. Generally, the active learning approaches follow the idea of uncertainty sampling [], wherein samples on which the current classifier is uncertain are selected to be trained. The distance from the hyperplane of the classifiers is always used as the uncertainty in previous work. However, this is hard to extend to multi-class classification due to the presence of multiple hyperplanes. Meanwhile, the design of active learning approach is usually based on binary class, so the label information is only used 1 or -1. In this way, it is hard to balance the distance to the multiple hyperplanes. To conquer this challenge, in the proposed method, for an instance, it is gave a set of labels whose length is equal to the number of classes as the multi-label classification. But it is different to the multi-label, in the multi-label classification, the set of labels of an instance may exist several labels are equal to 1. In our proposed multi-class active learning, the set of labels for an instance is just one label of them is equal to 1. A toy example is shown in fig.2. The label information will be sufficiently used.

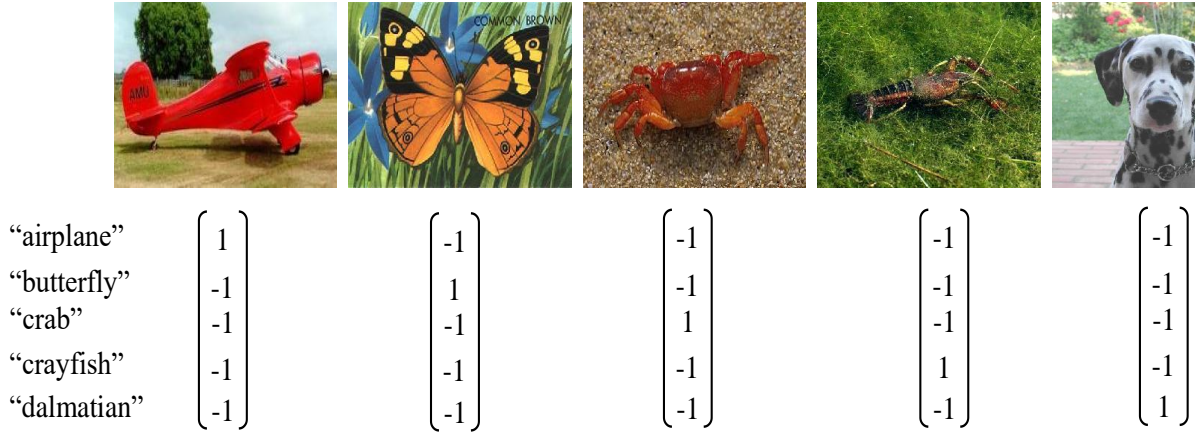

Fig.2 A toy example of the label definition of an instance for the proposed multi-class active learning method

Suppose we are given a data set $\mathcal{D} = \{x_1, x_2, \ldots, x_n\} \in R^d$, and it is randomly split into two data sets, the unlabeled data $\mathcal{D}_u = \{x_1, x_2, \ldots, x_u\}$ and the labeled data $\mathcal{D}_l = \{(x_1, y_1), (x_2, y_2), \ldots, (x_l, y_l)\}$, each $x_i$ in $\mathcal{D}_u$ and $\mathcal{D}_l$ is a $d$ dimensional feature vector. $\mathcal{D}_u$ is the candidate set and $\mathcal{D}_l$ is initially empty or with little labeled samples for active learning. The $y_i$ that corresponds to $x_i$ in $\mathcal{D}_l$ is denoted as a set of labels which has been defined above since we focus on the multi-class problems. Mathematically, $y_i = [y_{ik}]_{c\times 1} = [y_{i1}, y_{i2}, \ldots, y_{ic}]^T$, $c$ is the number of classes. If $x_i$ belongs to the class $c$, $y_{ic}$ is equal to 1, else $y_{ic}$ is equal -1. In our active learning, we iteratively select one instance $x_s$ from the pool of unlabeled data $\mathcal{D}_u$ to query its label and add the query instance to the labeled data $\mathcal{D}_l$. The goal is to improve the accurate model gradually. The symbols defined above will be used in our following discussion.

A. Informative information by Minimum Margin

In order to motivate the empirical risk for active learning based on margin, we first review the margin-based active learning. Mathematically, let $f^*$ be the classification model trained by the labeled samples, it can be expressed as:

$$f^* = \arg\min \sum_{i=1}^{l} \ell\left(Z_i, f(x_i)\right) + \lambda \|f\|_{\mathcal{H}}^2 \tag{4}$$

where $\mathcal{H}$ is RHKS endowed with kernel function $k(\cdot,\cdot): R^d \times R^d \to R$, $\ell(Z, f(x))$ is the loss function, $Z \in \{1, -1\}$. Given the classifier $f^*$, the margin-based approach choose the unlabeled sample which is the most uncertain for current classifier in $\mathcal{D}_u$

$$x_s^* = \arg\min_{x \in \mathcal{D}_u} |f^*(x)| \tag{5}$$

Proposition 1 shows the equivalent form of above formula in active learning setting.

**Proposition 1**: *The uncertain instance in Eq.(5) can be rewritten as*

$$x_s^* = \arg\max_{|Z_s|=1} \min_{f \in \mathcal{H}} \sum_{i=1}^{l} \ell\left(Z_i, f(x_i)\right) + \ell\left(Z_s, f(x_s)\right) + \lambda \|f\|_{\mathcal{H}}^2 \tag{6}$$

**Proof**: Define the object function by $\mathcal{U}(f)$

$$\mathcal{U}(f) = \sum_{i=1}^{l} \ell\left(Z_i, f(x_i)\right) + \lambda \|f\|_{\mathcal{H}}^2$$

the process of proof as follows

$$\begin{aligned}
x_s^* &= \arg\min_{x_s \in \mathcal{D}_u} |f^*(x_s)| \\
&= \arg\min_{x_s \in \mathcal{D}_u} \min_{f \in \mathcal{H}: \mathcal{U}(f) \le \mathcal{U}(f^*)} |f(x_s)| \\
&= \arg\min_{x_s \in \mathcal{D}_u, f \in \mathcal{H}: \mathcal{U}(f) \le \mathcal{U}(f^*)} |f(x_s)| \\
&= \arg\min_{x_s \in \mathcal{D}_u, f \in \mathcal{H}} |f(x_s)| + \Gamma \mathcal{U}(f) \\
&= \arg\max_{|Z_s|=1} \min_{x_s \in \mathcal{D}_u, f \in \mathcal{H}} \ell\left(Z_s - f(x_s)\right) + \Gamma \mathcal{U}(f)
\end{aligned}$$

Let $\Gamma = 1$, the Eq.(5) is same to the Eq.(6), it is well to demonstrate that they are equivalent.

we extend the binary margin-based approach to the multi-class problem, we first consider the simple case of active learning that does not take into account the label correlation. For each label, we learn a classifier independently. The object function of multi-class learning task with least square can be represented as:

$$\max_{|\hat{y}_s|=1^c} \min_{f_k \in \mathcal{H}, x_s} \sum_{i=1}^{l} \sum_{k=1}^{c} \left(y_{ik} - f_k(x_i)\right)^2 + \sum_{k=1}^{c} \left(\hat{y}_{sk} - f_k(x_s)\right)^2 + \lambda \sum_{k=1}^{c} \|f_k\|_{\mathcal{H}}^2 \tag{7}$$

where $f_k$ is the classifier of the $k^{th}$ class. $y_{ik}$ is the true label of instance $x_i$ corresponding to the label $k$, $y_{sk}$ is the pseudo label of the query sample $x_s$ corresponding to label $k$. $1^c$ is a vector of length $c$, with all entries 1. If we solve the formula (7) with regard to $\hat{y}_s$ with fixed f and $x_s$, we minimize the worst-case risk introduced by the query sample. In this case, the pseudo label $\hat{y}_{sk} = -\text{sign}(f_k(x_s)), k = 1, 2, \ldots, c$, therefore, the risk term becomes

$$\min_{f_k \in \mathcal{H}, x_s} \sum_{i=1}^{l} \sum_{k=1}^{c} \left(y_{ik} - f_k(x_i)\right)^2 + \sum_{k=1}^{c} \left(\hat{y}_{sk}^2 + 2|f_k(x_s)|\right) + \lambda \sum_{k=1}^{c} \|f_k\|_{\mathcal{H}}^2 \tag{8}$$

which is still the upper bound of the true risk. Conveniently, we use the linear model in the kernel space as the classifier, whose form is $f(x) = w^T \phi(x)$, $\phi(x)$ is feature mapping function as described above. The informative information we can measure by the objective function

$$\min_{w, x_s} \sum_{i=1}^{l} \sum_{k=1}^{c} \left( y_{ik} - w_k^T \phi(x_i) \right)^2 + \sum_{k=1}^{c} \left( \left| w_k^T \phi(x_s) \right|^2 + 2 \left| w_k^T \phi(x_s) \right| \right) + \lambda \sum_{k=1}^{c} \| w_k \|_{\mathcal{H}}^2 \quad (9)$$

Let $w = [w_1, w_2, \ldots, w_c]$ be the coefficient matrix. The above formula can be rewritten as

$$\min_{w, x_s} \sum_{i=1}^{l} \| y_i - w^T \phi(x_i) \|_2^2 + \| w^T \phi(x_s) \|_2^2 + 2 \left| w^T \phi(x_s) \right| + \lambda \| w \|_F^2 \quad (10)$$

Using a kernel form $w_k = \sum_{x_j \in T} \theta_{jk} \phi(x_j)$, and let $\theta_c = [\theta_{ik}]_{l \times 1} = [\theta_{1c}, \theta_{2c}, \ldots, \theta_{lc}]^T$, $\phi(T) = [\phi(x_1), \phi(x_2), \ldots, \phi(x_l)]^T$, $x_k \in T$, therefore, $w_k = \theta_k^T \phi(T)$, and the coefficient matrix can be reformed by $\theta$:

$$w = \left[ \theta_1^T \phi(T), \theta_2^T \phi(T), \ldots, \theta_c^T \phi(T) \right] = \theta^T \phi(T)$$

where $\theta = [\theta_1, \theta_2, \ldots, \theta_c]$ is the coefficient matrix of $w$ in kernel space. Define $Y = [y_k]_{l \times c} = [y_1, y_2, \ldots, y_l]^T$ is the label matrix, $R \in R_+^{c \times c}$ is an incidence matrix between labels. Meanwhile, a function $vec(\cdot)$ is introduced to convert a matrix into a vector along the column. Hence, the objective function is modified by taking into account the label correlation

$$\min_{\theta} \left\| vec(Y) - (vec(\theta)(R \otimes K_{\mathcal{D}_l \mathcal{D}_l})) \right\|_2^2 + \left\| vec(\theta)^T (R \otimes K_{\mathcal{D}_l s}) \right\|_2^2 + 2 \left| vec(\theta)^T (R \otimes K_{\mathcal{D}_l s}) \right| \\ + \lambda vec(\theta)^T (R \otimes K_{\mathcal{D}_l \mathcal{D}_l}) vec(\theta) \quad (11)$$

where $\otimes$ is the kronecker product between matrixes, $K$ is the kernel matrix with its elements $K_{ij} = k(x_i, x_j) = \phi(x_i)^T \phi(x_j)$.

B.  Representative information by Maximum Mean Discrepancy

In the proposed framework, the MMD is adopted to measure the representative information which is the evaluation of distribution difference between two sets of samples. In active learning, it can constrain the distribution of the labeled and query samples, and make it similar to the overall sample distribution as much as possible. According the description about the MMD above, the MMD can empirically calculate in active learning as

$$MMD_\phi \left( \mathcal{D}_l \cup x_s, \mathcal{D}_u / x_s \right) = \left\| \frac{1}{l+1} \sum_{x_i \in \mathcal{D}_l \cup x_s} \phi(x_i) - \frac{1}{u-1} \sum_{x_j \in \mathcal{D}_u / x_s} \phi(x_j) \right\| \quad (12)$$

Actually, $x_s$ is the target sample which is the goal of the proposed active learning method to query. So the above formula should be defined as an alternative representation, which can select $x_s$ from the unlabeled data set $\mathcal{D}_u$ by optimization solution. According [11], MMD can be transferred into the following representation

$$\min_{\alpha: \alpha_i \in \{0,1\}, \alpha^T \mathbf{1} = 1} \frac{1}{2} \alpha^T k_1 \alpha - \frac{l+1}{l+u} \mathbf{1}_l k_2 \alpha + \frac{u-1}{l+u} \mathbf{1}_u k_3 \alpha + const. \quad (13)$$

where $\alpha$ is an indicator vector of length $u$, note that if $x_s$ is selected in the unlabeled data set $\mathcal{D}_u$ then the corresponding $\alpha_s$ will be 1, otherwise $\alpha_i$ will be 0. **1** is a vector of the same to the dimension $\alpha$ with all entries 1. $\mathbf{1}_l$ and $\mathbf{1}_u$ are vectors of length $l$ and $u$ respectively, with all elements 1. The other terms are given as follows. K is the kernel matrix with its element described above. $k_1 = k_3 = K_{\mathcal{D}_u \mathcal{D}_u}$, $k_2 = K_{\mathcal{D}_l \mathcal{D}_u}$, $K_{AB}$ denotes the kernel matrix between A and B. Since the second term and the third have same variable, it can be simplified as a form of standard quadratic programming

$$\min_{\alpha: \alpha_i \in \{0,1\}, \alpha^T \mathbf{1} = 1} \frac{1}{2} \alpha^T K_1 \alpha + K_2 \alpha \tag{14}$$

where $K_1 = K_{\mathcal{D}_u \mathcal{D}_u}$, $K_2 = \frac{u-1}{l+u} \mathbf{1}_u k_3 - \frac{l+1}{l+u} \mathbf{1}_l k_2$.

C. The Hybrid Informative and Representative Multi-class Active Learning

Based on the discussion above, a hybrid informative and representative information multi-class active learning method (IR-AL) is proposed. The framework of the proposed method can be representation as follows

$$\min_{\substack{\theta \\ \alpha: \alpha_i \in \{0,1\}, \alpha^T \mathbf{1}=1}} \left\| vec(Y) - \left( vec(\theta)(R \otimes K_{\mathcal{D}_l \mathcal{D}_l}) \right)^T \right\|_2^2 + \left\| vec(\theta)^T (R \otimes K_{\mathcal{D}_l s}) \right\|_2^2 + 2 \left| vec(\theta)^T (R \otimes K_{\mathcal{D}_l s}) \right|$$

$$+ \lambda vec(\theta)^T (R \otimes K_{\mathcal{D}_l \mathcal{D}_l}) vec(\theta) + \beta \left( \frac{1}{2} \alpha^T K_1 \alpha + K_2 \alpha \right) \tag{15}$$

where $\beta$ is the trade-off weight to balance the informative and representative information. According the discussion about measuring the informative and representative information, respectively, the empirical risk of the object function is described above approximates to the true risk upper bound under the original distribution. It is not difficult to imagine that the query sample $x_s$ is the bond between the two parts. So the $x_s$ in informative part is also can be selected with the indicator vector $\alpha$ from unlabeled data set $\mathcal{D}_u$. Hence, the above formula can be reformed as

$$\min_{\substack{\theta \\ \alpha: \alpha_i \in \{0,1\}, \alpha^T \mathbf{1}=1}} \left\| vec(Y) - \left( vec(\theta)(R \otimes K_{\mathcal{D}_l \mathcal{D}_l}) \right)^T \right\|_2^2 + \sum_{i=1}^u \alpha_i \left( \left\| vec(\theta)^T (R \otimes K_{\mathcal{D}_l i}) \right\|_2^2 + 2 \left| vec(\theta)^T (R \otimes K_{\mathcal{D}_l i}) \right| \right)$$

$$+ \lambda vec(\theta)^T (R \otimes K_{\mathcal{D}_l \mathcal{D}_l}) vec(\theta) + \beta \left( \frac{1}{2} \alpha^T K_1 \alpha + K_2 \alpha \right) \tag{16}$$

Intuitively, the above problem is not convex. In this way, the alternating optimization is adopted [28]. If $\alpha$ is fixed, the representative term is a constant, the above objective is a problem which is to find the best classifier with the labeled data set $\mathcal{D}_l$ and the query sample $x_s$.

$$\min_{\theta} \left\| vec(Y) - \left( vec(\theta)(R \otimes K_{\mathcal{D}_l \mathcal{D}_l}) \right)^T \right\|_2^2 + \left\| vec(\theta)^T (R \otimes K_{\mathcal{D}_l s}) \right\|_2^2 + 2 \left| vec(\theta)^T (R \otimes K_{\mathcal{D}_l s}) \right|$$

$$+ \lambda vec(\theta)^T (R \otimes K_{\mathcal{D}_l \mathcal{D}_l}) vec(\theta) + const \tag{17}$$

The above formula can be solved by the alternating direction method of multipliers (ADMM) [29]. If $\theta$ is fixed, the terms in informative part will be constant except the term which contains the indicator $\alpha_k$. The objective becomes

$$\min_{\alpha:\alpha_i \in \{0,1\}, \alpha^T \mathbf{1}=1} \sum_{i=1}^{u} \alpha_i \left( \left\| vec(\theta)^T (R \otimes K_{\mathcal{D}_i}) \right\|_2^2 + 2 \left| vec(\theta)^T (R \otimes K_{\mathcal{D}_i}) \right| \right) + \beta \left( \frac{1}{2} \alpha^T K_1 \alpha + K_2 \alpha \right) + const \tag{18}$$

Here, we define $K_3(i) = \left\| vec(\theta)^T (R \otimes K_{\mathcal{D}_i}) \right\|_2^2 + 2 \left| vec(\theta)^T (R \otimes K_{\mathcal{D}_i}) \right|$, the above objective function can be rewritten as

$$\alpha^* = \min_{\alpha:\alpha_i \in \{0,1\}, \alpha^T \mathbf{1}=1} \frac{\beta}{2} \alpha^T K_1 \alpha + (\beta K_2 + K_3) \alpha + const \tag{19}$$

This is a standard quadratic programming (QP) form to the indicator vector $\alpha$. If the constraint condition $\alpha : \alpha_i \in \{0,1\}, \alpha \mathbf{1}^u = 1$ is restrict for the QP, the time cost will be expensive to find the best value for the QP. So we relax the $\alpha_k$ from 0 to 1, the value in $\alpha^*$ which is closest to 1 will be set as 1, otherwise, it will be set as 0.

D. The Solution

In this section, we will discuss the process to solve our proposed method in details. According [28], the alternating optimization is to solve a non-convex problem by changing one kind of variables and fixing the others, and this procedure is alternating for all variables until satisfy the convergence condition. Hence, two steps are included to solve (16) by alternating optimization. As described above, firstly, if $\alpha$ is fixed, the Eq.(17) can be solved by ADMM to calculate the best $\theta$. Secondly, $\theta$ is fixed which is calculated in the first step, the Eq.(18) is solved by employing the quadratic programming (QP) to solve $\alpha$.

Firstly, the indicator vector $\alpha$ is fixed, the objective (16) becomes (17), for the computational simplicity, we define the incidence matrix R is an identity matrix and an auxiliary variable $a = vec(\theta)^T (R \otimes K_{\mathcal{D}_s})$ is introduced. The objective function (17) becomes

$$\min_{\theta} \left\| vec(Y) - \left( vec(\theta)(R \otimes K_{\mathcal{D}_l \mathcal{D}_l}) \right)^T \right\|_2^2 + \|a\|_2^2 + 2|a| + \lambda vec(\theta)^T (R \otimes K_{\mathcal{D}_l \mathcal{D}_l}) vec(\theta) + const \tag{20}$$

$$s.t. \quad a - vec(\theta)^T (R \otimes K_{\mathcal{D}_s}) = \mathbf{0}, \forall x_s \in \mathcal{D}_u$$

where $\mathbf{0}$ is a vector of length $c$, with all entries 0. To construct the augmented Lagrangian of Eq.(20) as

$$L_\rho = \left\| vec(Y) - \left( vec(\theta)(R \otimes K_{\mathcal{D}_l \mathcal{D}_l}) \right)^T \right\|_2^2 + \|a\|_2^2 + 2|a| + \lambda vec(\theta)^T (R \otimes K_{\mathcal{D}_l \mathcal{D}_l}) vec(\theta)$$

$$+ \left( a - vec(\theta)^T (R \otimes K_{\mathcal{D}_s}) \right) \xi^T + \frac{\rho}{2} \left\| a - vec(\theta)^T (R \otimes K_{\mathcal{D}_s}) \right\|_2^2$$

According [29], the updating rules can be obtained as follows

$\theta^{k+1} = B^{-1} b^T$ with $B = (R \otimes K_{\mathcal{D}_l \mathcal{D}_l})^2 + \frac{\rho}{2} (R \otimes K_{\mathcal{D}_s})(R \otimes K_{\mathcal{D}_s}^T) + \lambda (R \otimes K_{\mathcal{D}_l \mathcal{D}_l})$, and

$b = vec(Y_L)(R \otimes K_{\mathcal{D}_l \mathcal{D}_l}) + \frac{1}{2} \xi^k (R \otimes K_{\mathcal{D}_s}^T) + \frac{\rho}{2} a^k (R \otimes K_{\mathcal{D}_s}^T)$

$a^{k+1} = sign(v)(|v| - \omega)_+$ with $v = \dfrac{\rho (\theta^{k+1})^T (R \otimes K_{\mathcal{D}_s}) - \xi^k}{\rho + 2}$, $\omega = \dfrac{2}{\rho + 2}$, $(q)_+ = \max(0, q)$

$\xi^{k+1} = \xi^k + \rho \left( a^{k+1} - (\theta^{k+1})^T (R \otimes K_{\mathcal{D}_s}) \right)$

By employing AMDD to solve $\theta$ in the first step, therefore, in the second step, $\theta$ is foregone, the objective becomes Eq.(19), it is a standard QP problem for $\alpha$. This problem can be solved using standard QP toolboxes such as MOSEK[1] or the function quadprog[2] in MATLAB. If the two steps are all convergence, for the compute $\alpha$, the unlabeled sample in $\mathcal{D}_u$ which is corresponding to the position of the largest one element in $\alpha$ is the query sample $x_s$.

## IV. EXPERIMENTS

To investigate the proposed method performance, in our experiments, we compare the proposed method with the following three state-of –the-art active learning methods.

- QUIRE: min-max based active learning [10], an approach that queries both informative and representative information instances.
- Adaptive: active learning that combining the uncertainty information and representative information together with product [16], so this is an approach to query informative and representative instances.
- MP-AL: active learning based on marginal probability distribution matching [11], an approach that prefers representative instances.

### A. Settings

In our experiments, twelve data sets are used from UCI benchmarks in our study. The characteristics of these data set are summarized in Table I. *vote*, *ionosphere*, *image* and *australian* are benchmark data sets with binary class. Since many active learning methods are designed on binary class, in order to demonstrate the effectiveness about the proposed method, we conduct our experiments on these four benchmark binary class data sets. *balance, iris, vehicle, wine, waveform, vowel, glass,* and *landsat* are multi-class data sets with more than two classes. We randomly divide each data set into two parts with percentage 60% and 40%, the part of 40% is used as the test data and the other part is used as the unlabeled data for active learning. We assume that no labeled data is available at the very beginning of active learning. For Adaptive method, the initial labeled data are needed. The initial labeled data randomly are sampled from the 60% data with samples from each class. For each class, there is just one sample labeled and added to the initial labeled set, therefore, the number of the labeled data is only enough to train an initial classifier. At each iteration, an instance is selected to solicit its label and the classification model is retrained. Meanwhile, the retrained classification model is evaluated by its performance on the test data. Since active learning is an iterative process, we stop our experiments on each data set when the learning accuracy does not increase for any method. But for some data sets, the experiments stop much earlier due to the limited samples, i.e. *vowel, ionosphere*. For each data set, we run 10 times with different randomly partition of the data set. And for the compared methods, the parameters we used are according the original papers. In all experiments, for fairness, the LIBSVM [30] is used to train a classifier in all methods, the classification accuracy curve of the SVM classifier after each query are used for evaluation metrics. A RBF kernel is used in all kernel calculation. For the SVM classifier, the main parameters are the penalty coefficient $C$ and the kernel bandwidth $g$. We set the two parameters with empirical value with 100 and $1/d$, respectively, where $d$ is the dimension of the original data. In our proposed method, there are also two parameters existing, the regularization weight $\lambda$ and the trade-off parameter $\beta$. We set the regularization weight $\lambda = 0.1$. The trade-off parameter $\beta$ is chose from a candidate set {1, 2, 10, 100, 1000} by cross validation. In the experiments, a QP problem need to be solved, we adopt the MOSEK toolbox in our experiments as the solver. For the Adaptive and QUIRE method, the inverse of the kernel matrix should be calculated, hence, if the data set is too large, when running the two methods, it requests the memory of computer high, but we also want to keep the performance of these methods, so in our experiments, the relatively small benchmark data sets are selected.

Table I: Characteristics of the data sets, including the numbers of the corresponding features and samples.

| Dataset | #Feature | #Instance |
|---------|----------|-----------|
| balance | 4 | 625 |
| iris | 4 | 150 |
| vehicle | 18 | 846 |
| wine | 13 | 178 |
| waveform | 21 | 5000 |
| vowel | 10 | 528 |
| australian | 14 | 690 |
| glass | 9 | 214 |
| ionosphere | 34 | 351 |
| image | 18 | 2086 |
| vote | 16 | 435 |
| landsat | 36 | 2000 |

B. Comparison with State-of-the-Art Methods

Fig.3 shows the classification accuracy of the proposed method and the compared methods with varied numbers of query samples. Table II shows the win/tie/loss counts of the IR-AL versus the other state-of-the-art methods.

Intuitively, we can take a glance at that the proposed method IR-AL performs better much better than the compared methods. Firstly, we observe that the performance of QUIRE is better than the other two compared methods. It works well on half number of the benchmark data sets, but performance poorly on the others. Since QUIRE is margin-based approach from the viewpoint of min-max, and it requires the unlabeled data in the semisupervised learning for boosting the learning performance, hence, we attribute the poorly performance on some data sets of the QUIRE to the fact that the unlabeled data structure does not satisfy the semi-supervised assumptions. The behavior of the Adaptive method is performance worst in all the methods. Adaptive method is a method to combine the uncertainty and representativeness with product, but they are calculated respectively. Meanwhile, the representative information is measured by the Gaussian Process. Therefore, it needs a large amount of unlabeled data to evaluate the distribution, if the unlabeled data are not enough to evaluate the distribution correctly, the performance will be bad. Since Adaptive needs to calculate an inverse matrix, so the data sets we use are not large. This may be the reasonable analysis that Adaptive does not yield good performance on most data sets since the number of unlabeled data is not large enough. MP-AL performs better than Adaptive method, although it is just use the MMD as the query criterion. It just measures the representativeness of the query instance, so it also need large amount of data. The reason that it performs better than Adaptive method may be that it is no requirement about the unlabeled data distribution.

Finally, we observe that our method seldom performs worse than the compared methods. According fig.3 and table II, the proposed method of choosing informative and representative instances is successfully for the multiple class data sets. These results demonstrate that both the informative information and representative information are significant to design a good active learning approach. If a proper trade-off weight is used, it will boost the active learning performance.

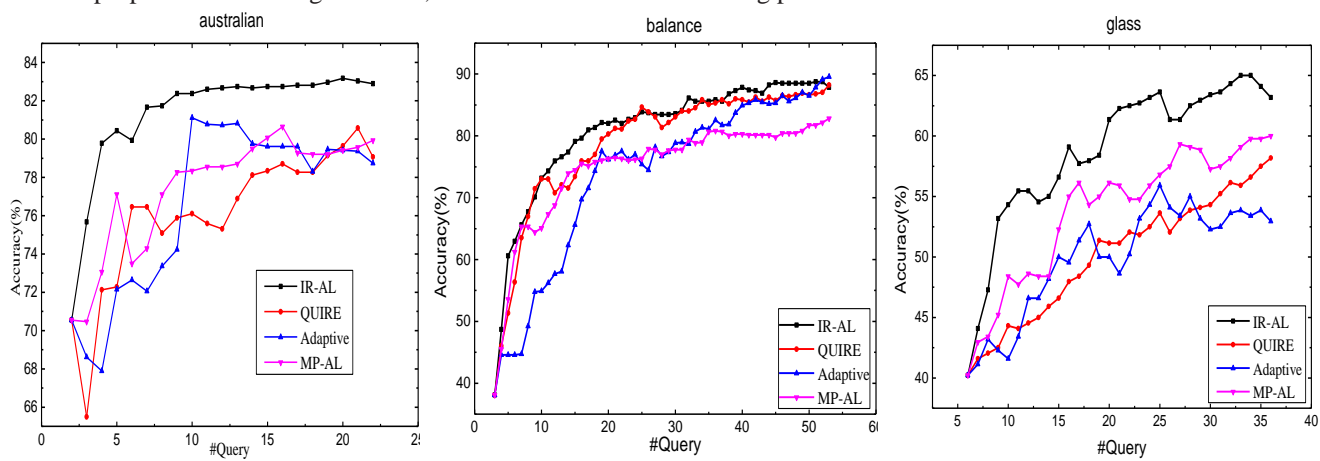

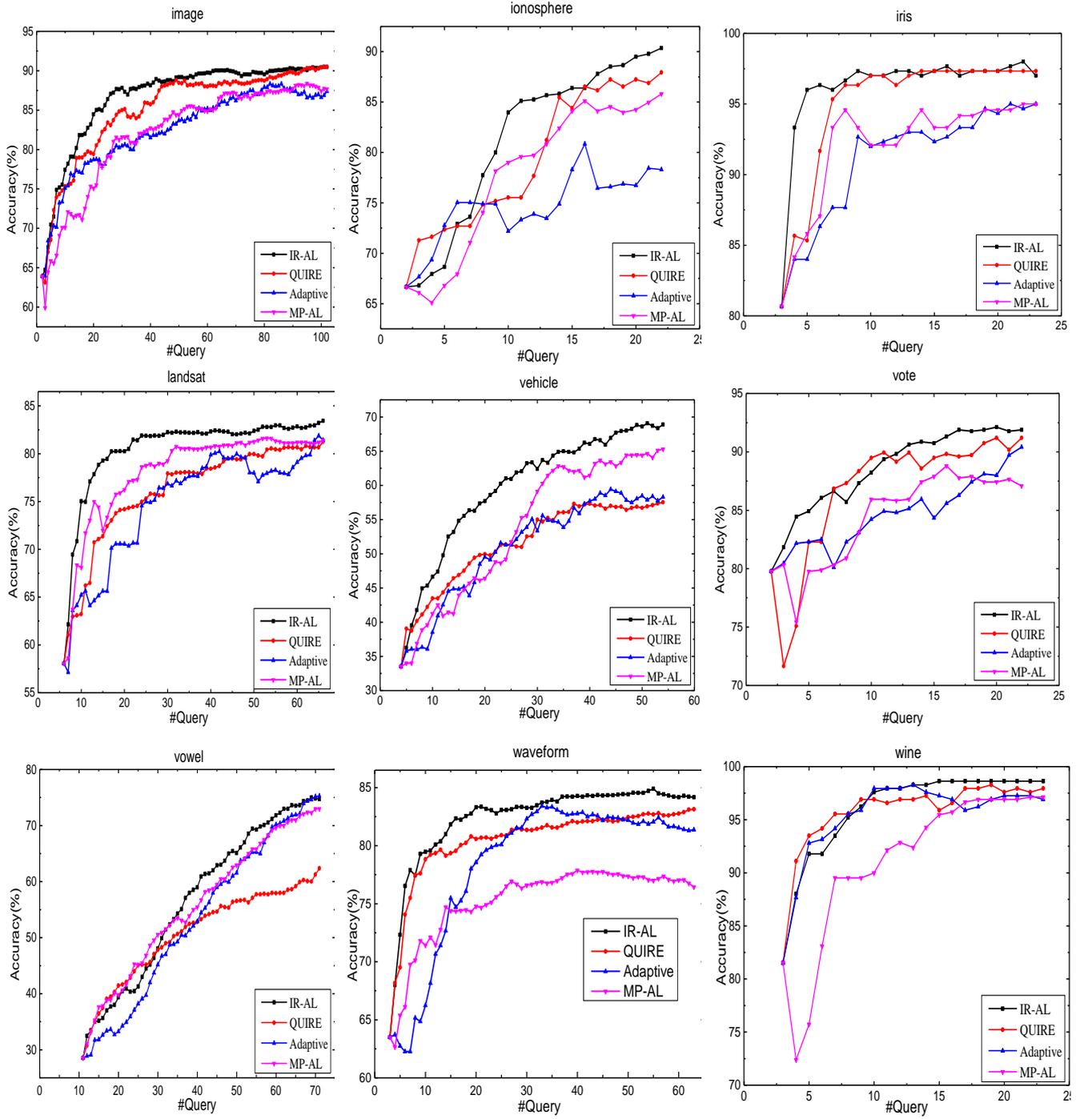

Fig.3 Comparison of different active learning methods on twelve UCI benchmark data sets. The curve shows the learning accuracy over queries, and each curve represents the average result of 5 runs.

TABLE II: Win/Tie/Loss counts of the proposed method versus the other state-of-the-art methods Based on Paired *t*-Tested at 95 Percent Significance Level

| Dataset | Vs QUIRE | Vs Adaptive | Vs MP-AL |
|---|---|---|---|
| image | 15/5/5 | 25/0/0 | 25/0/0 |
| ionosphere | 17/8/0 | 25/0/0 | 24/1/0 |
| iris | 7/18/0 | 25/0/0 | 15/5/0 |
| landsat | 25/0/0 | 23/2/0 | 21/4/0 |
| vehicle | 25/0/0 | 25/0/0 | 24/1/0 |
| vote | 11/11/3 | 17/6/2 | 20/2/3 |
| vowel | 20/5/0 | 19/2/4 | 14/3/8 |
| waveform | 17/6/2 | 25/0/0 | 25/0/0 |
| wine | 5/17/3 | 3/22/0 | 20/5/0 |
| australian | 13/12/0 | 17/8/0 | 16/9/0 |
| balance | 13/9/3 | 23/2/0 | 24/1/0 |
| glass | 15/7/3 | 21/4/0 | 19/1/5 |

C. Sensitivity Analysis

In our proposed method, the parameter β which balances the informative information and representative information in the formula is single factor that influences experimental results. In our experiments, we obtain the value of β from a candidate set {1, 2, 10, 100, 1000} in every runs for our proposed method. To test the influence of this parameter, we show the results on a UCI benchmark data set: semeion handwritten digit which contains 1593 instances with 256 features. The experimental settings are same to the foregoing experiments except the parameter β. We report our results in Fig.3. From the results, we directly observe that at the beginning of the curves, there is a poor performance on all cures, so we can consider that this is no immediate relationship with the parameter. Ignoring this phenomenon, we can observe that our method is more sensitive to the trade-off parameter β. When we use the candidate set by cross validation to choose the best value of β, the curve is always higher than the cures with fixed β. When β has a small value such as {1, 2, 10}, there is no obvious changes for the three curves. When β is increasing, the performance also will be better. However, in a certain interval, the curves changes not clearly. Although the changes is not clearly in a certain interval, the larger values gets a better performance. Therefore, a larger value of parameter β is recommended. In this way, much more attention is paid on data distribution, which can boost the active leaning performance.

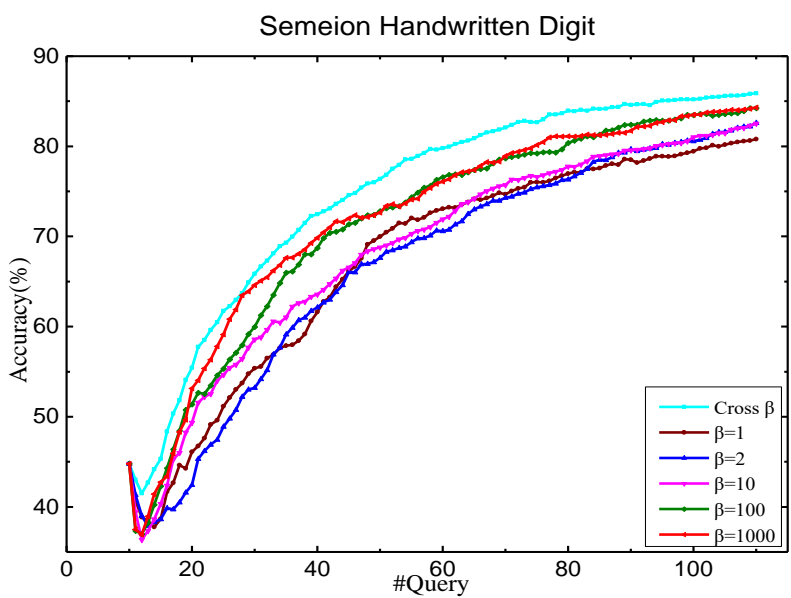

Fig.3 Performance comparison using different trade-off parameters on semeion handwritten digit data set for the proposed method. Each curve represents the average result of 10 runs.

V. CONCLUSION

In this paper, we generalize the empirical risk principle to the active setting and proposed a novel multi-class active learning approach. By combing the informative part and representative part together, the proposed method can reduce the empirical risk rapidly since the data distribution is preserved. Meanwhile, we make full use of the label information, for each instance, it owns a set of labels, which just has a positive label. In this situation, our proposed method will be more practical since most problems in real world are multiple classes, and it also will boost the active learning performance, because it makes the instance that we query is closest to all the classification hyperplane of binary class. The superior performance of our proposed method is demonstrated by the experiments that we conduct on twelve UCI benchmark data sets. Compared with the state-of-the-art methods, there is a limit for our method. It is the trade-off parameter to balance the informative information and representative information. Since the experimental results are sensitive to the parameter, this is a critical problem to restrict the applications of our method. Our future work is to obtain the value of parameter adaptively, which can make our method more practical, and try to develop our method to multi-label classification and semi-supervised learning, which can expand the application of our work.